\documentclass[letterpaper, 10 pt, conference]{ieeeconf}  

\IEEEoverridecommandlockouts                              

\usepackage{url}
\usepackage{times}
\usepackage{epsfig}
\usepackage{graphicx}
\usepackage{amsmath}
\usepackage{amssymb}
\usepackage{booktabs}
\usepackage[skip=0pt]{caption} 
\usepackage{chapterbib}
\usepackage{subcaption}
\usepackage{hyperref}
\usepackage{textpos}

\title{\LARGE \bf
SymbioLCD: Ensemble-Based Loop Closure Detection using CNN-Extracted Objects and Visual Bag-of-Words
}

\author{Jonathan J.Y. Kim$^{1,2*}$, Martin Urschler$^{1}$, Patricia J. Riddle$^{1}$, and J\"org S. Wicker$^{1}$ \\ $^{1}$School of Computer Science, University of Auckland, New Zealand  \:\:$^{2}$Callaghan Innovation, New Zealand
\\ {\tt\small jkim072@aucklanduni.ac.nz}, {\tt\small \{martin.urschler,p.riddle,j.wicker\}@auckland.ac.nz}
\thanks{*\textit{This research was supported by Callaghan Innovation, New Zealand's Innovation Agency}}
}

\newcommand{\etal}{{\em et al.}}

\begin{document}

\maketitle
\thispagestyle{empty}
\pagestyle{empty}

\begin{textblock*}{180mm}(.01\textwidth,-6.8cm)
Accepted at IROS 2021.
© 2021 IEEE.  Personal use of this material is permitted.  Permission from IEEE must be obtained for all other uses, in any current or future media, including reprinting/republishing this material for advertising or promotional purposes, creating new collective works, for resale or redistribution to servers or lists, or reuse of any copyrighted component of this work in other works.
\end{textblock*}


\begin{abstract}

Loop closure detection is an essential tool of Simultaneous Localization and Mapping (SLAM) to minimize drift in its localization. Many state-of-the-art loop closure detection (LCD) algorithms use visual Bag-of-Words (vBoW), which is robust against partial occlusions in a scene but cannot perceive the semantics or spatial relationships between feature points. CNN object extraction can address those issues, by providing semantic labels and spatial relationships between objects in a scene. Previous work has mainly focused on replacing vBoW with CNN derived features.  
In this paper we propose SymbioLCD, a novel ensemble-based LCD that utilizes both CNN-extracted objects and vBoW features for LCD candidate prediction. When used in tandem, the added elements of object semantics and spatial-awareness creates a more robust and symbiotic loop closure detection system. The proposed SymbioLCD uses scale-invariant spatial and semantic matching, Hausdorff distance with temporal constraints, and a Random Forest that utilizes combined information from both CNN-extracted objects and vBoW features for predicting accurate loop closure candidates. Evaluation of the proposed method shows it outperforms other Machine Learning (ML) algorithms - such as SVM, Decision Tree and Neural Network, and demonstrates that there is a strong symbiosis between CNN-extracted object information and vBoW features which assists accurate LCD candidate prediction. Furthermore, it is able to perceive loop closure candidates earlier than state-of-the-art SLAM algorithms, utilizing added spatial and semantic information from CNN-extracted objects.

\end{abstract}


\section{Introduction}
\label{sec:introduction}
Simultaneous Localisation and Mapping (SLAM) has been widely used in the field of robotics, for simultaneously building a 3D map and finding the pose of a camera.
Indirect or feature-based Simultaneous Localisation and Mapping relies on feature extraction for localization and mapping of a scene. Compared to direct SLAM \cite{elasticfusion, LSDSLAM} which uses direct pixel intensities or edges \cite{reslam}, indirect SLAM uses a sparse set of feature points which allows easier transition from images to geometry and is robust to partial occlusion \cite{feature_based, dense_map}. However, over time small drifts will start to occur due to errors in motion estimation and feature extraction, and accumulation of these small errors in each frame will cause a large drift by the end of the trajectory. The uncertainty in the map will continue to grow until a loop closure has been detected in the map. SLAM relies on Loop Closure Detection (LCD) to remove these drifts using bundle adjustment, by returning to where the trajectory has started and re-observing features it has mapped earlier \cite{DBOW2}. 

\begin{figure}[t]
\begin{center}
\includegraphics[width=0.8\linewidth]{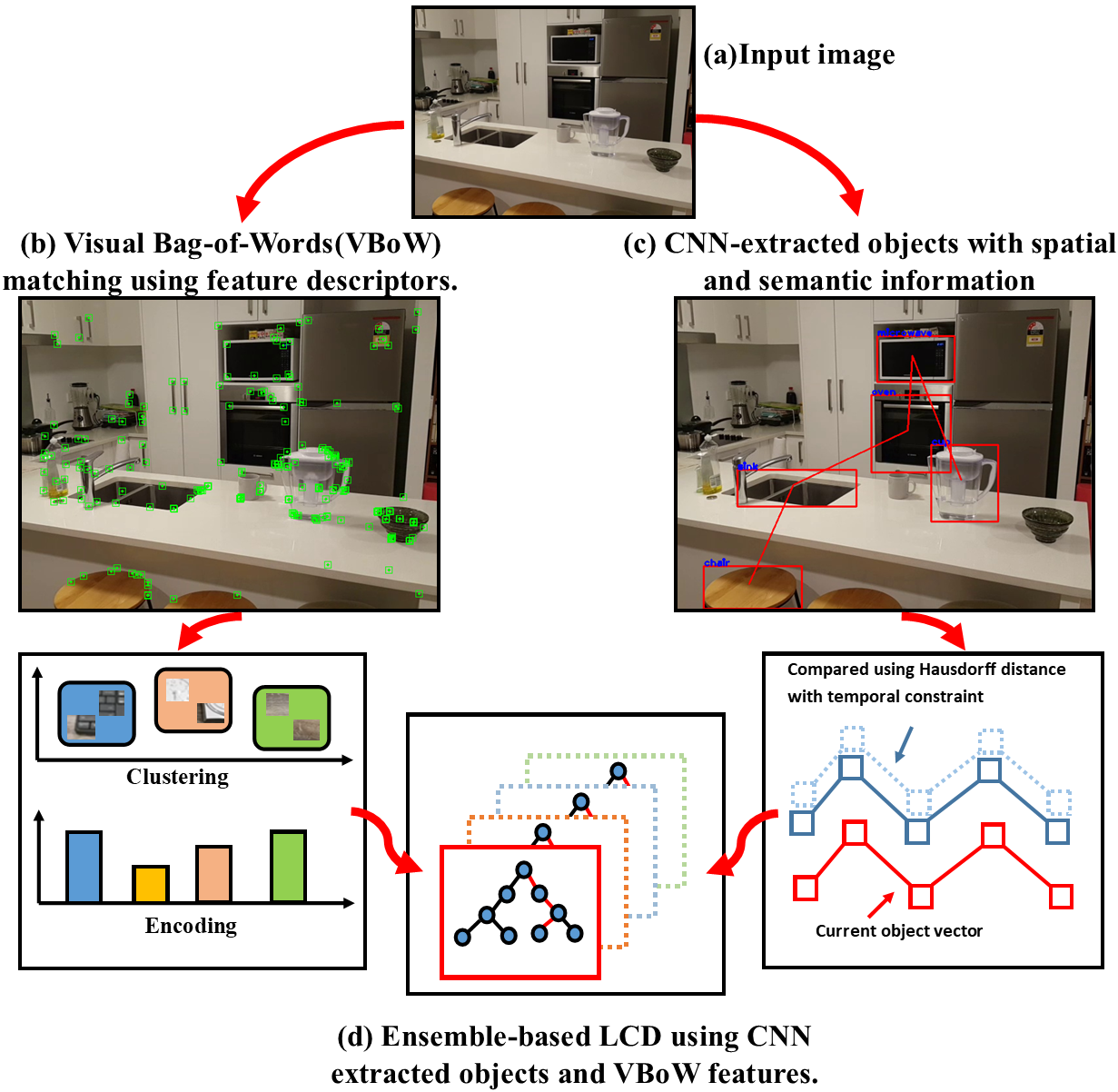}
\end{center}
\caption{Basic overview of SymbioLCD}
\label{fig:title}
\end{figure}

Although it is a crucial element, LCD can be laborious, since it is difficult to know if the features observed in the current frame are the same ones that were observed when the trajectory started.
Many state-of-the-art SLAM systems, such as ORB-SLAM2 \cite{ORBSLAM}, rely on visual Bag-of-Words (vBoW) for loop closure detection. vBoW finds its loop closure candidates by clustering and encoding the feature points found in a scene, then compares them against the vocabulary stored in the vBoW dictionary. If the match occurrence is above a certain threshold, the loop is closed to minimise the drift error~\cite{BagOfWords}. Since vBoW utilizes sparse feature points, such as SURF~\cite{SURF}, BRIEF~\cite{BRIEF} and ORB~\cite{ORB}, it is fast and robust against partial occlusions - making it a popular LCD component for many state-of-the-art SLAM systems.
However, it has a major disadvantage - during the clustering and encoding of feature points, it loses all the spatial relationships between feature points in the scene.

Convolutional Neural Networks (CNN) have recently made a leap in many applications such as object segmentation and instance classification. State-of-the-art CNN object detectors, such as Mask R-CNN, have shown great results in finding instances of objects even in complex and dynamic environments \cite{maskrcnn}. There has been some recent work done in the field to use instances of CNN-extracted (CNN-e) objects rather than feature points for LCD \cite{Saliency} showing good accuracy against state-of-the-art algorithms.
We propose to use CNN-e objects, because they provide two distinctive advantages. First, it is easier to recognise the scene as it provides semantics of objects, i.e. cups, books and chairs, rather than just selected features of objects, reducing the ambiguity of knowing what objects are in the scene and which feature points belong to which objects. Second, it allows LCD to utilize spatial relationships between objects, such as their placement in the scene and the relative distances between them.

\begin{figure*}[ht]
\begin{center}
\includegraphics[width=1.0\linewidth]{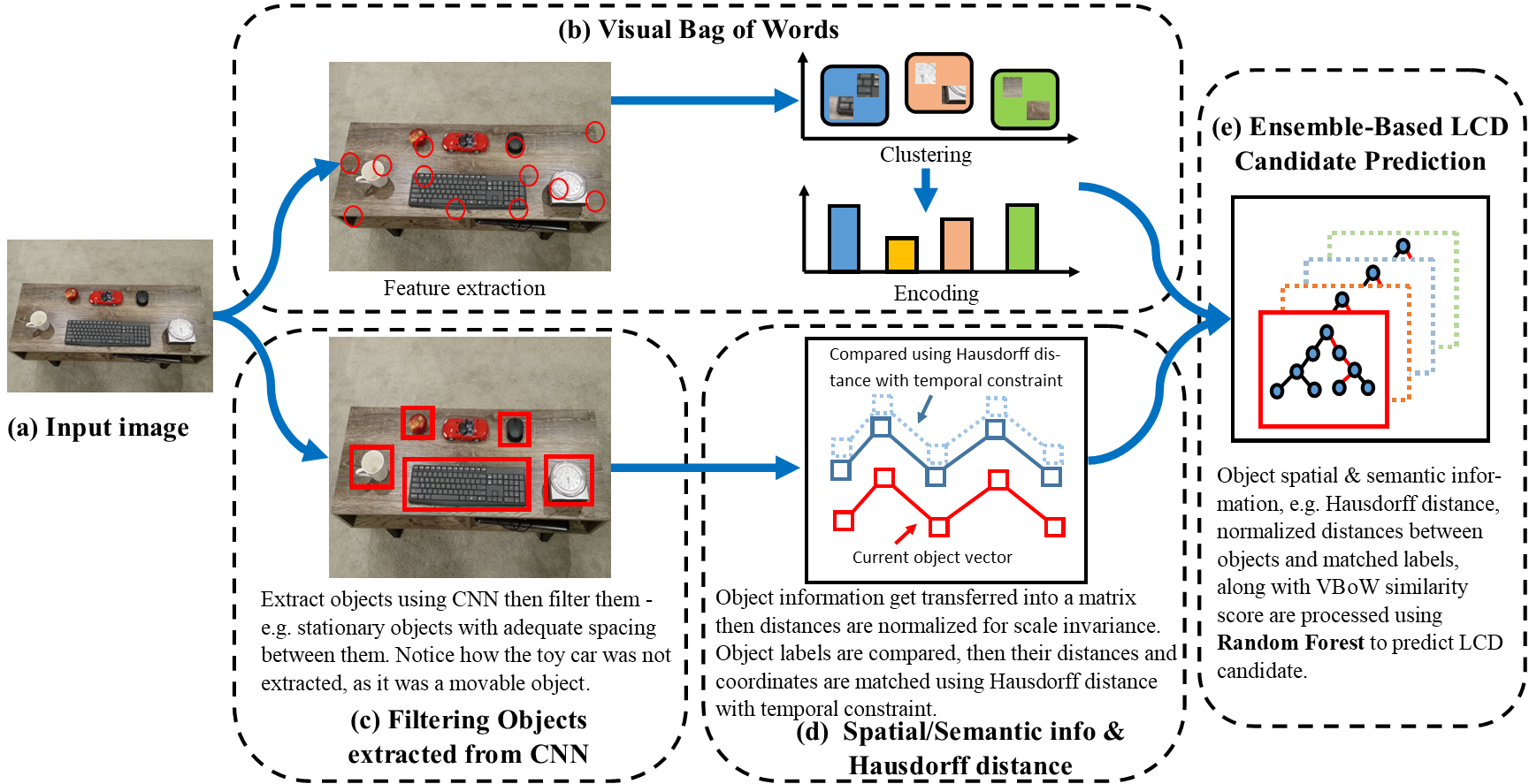}
\end{center}
   \caption{Detailed overview of SymbioLCD }
\label{fig:det_overview}
\end{figure*}

In this paper we present SymbioLCD, a novel ensemble-based loop closure detection system that combines the strength of both CNN-e objects and vBoW features symbiotically, to construct a more robust and complementary loop closure detection system.
In order to fully utilize both CNN-e object information and vBoW features, we propose to use Random Forest algorithm for LCD candidate prediction, which is an ensemble-based algorithm using multiple Decision Trees \cite{randforest}. Random Forest is faster to train compared to Neural Networks, and less prone to overfitting compared to other classical ML algorithms, making it an ideal candidate for the LCD predictor. Figure \ref{fig:title} shows the basic framework and Figure \ref{fig:det_overview} shows the detailed overview of SymbioLCD.

The main contributions of this paper are as follows: (a) A novel spatially aware ensemble-based LCD that uses both CNN-e objects and vBoW features symbiotically to create a more robust and complementary LCD system, (b) A scale-invariant semantic matching algorithm and a novel use of Hausdorff distance with temporal constraints, (c) Early detection of loop closure candidates utilizing both spatial \& semantic information and vBoW features, and (d) Three new datasets representing normal household areas with common objects - lounge, kitchen and garden, with varying trajectories. Our datasets are available at \url{https://doi.org/10.17608/k6.auckland.14958228}.

The paper is organized as follows: In Section \ref{sec:relatedwork} we cover relevant related work, Section \ref{sec:proposedmethod} describes the proposed methods, Section \ref{sec:experiments} presents experimental results, and Section \ref{sec:conclusion} presents the conclusion and future work.

\section{Related Work}
\label{sec:relatedwork}
In this section we review feature descriptors, how vBoW is utilized in indirect SLAM systems, recent trends in utilizing CNN derived descriptors in SLAM systems, and ensemble-based machine learning algorithms. 

\textbf{Feature descriptors} Indirect SLAM reduces an image into a sparse set of key points, which are matched using feature descriptors~\cite{Feature_matching}, such as SIFT~\cite{SIFT} and  SURF~\cite{SURF}, or binary descriptors such as BRIEF~\cite{BRIEF} and ORB~\cite{ORB}. Binary descriptors can be faster and more efficient, but traditional feature descriptors tend to have less uncertainty and better accuracy~\cite{Brisk_comp}.

\textbf{Visual Bag-of-Words} Many state-of-the-art indirect SLAM systems, such as ORB-SLAM2~\cite{ORBSLAM}, DynaSLAM~\cite{dynaslam} and LSD-SLAM~\cite{LSDSLAM} utilize vBoW to perform LCD. Two of the most popular vBoW based approaches are FAB-MAP~\cite{FABMAP} and DBoW2~\cite{DBOW2}. FAB-MAP uses a Chow-Liu tree to learn a generative model of appearances. The algorithm's complexity is linear, allowing fast online loop closure detection~\cite{FABMAP}.
DBoW2 uses a tree-structured dictionary created from offline training over a large dataset, and it compares new feature points against the dictionary to estimate the co-occurrence of features in the frame~\cite{DBOW2}. vBoW based algorithms cluster and encode feature points found in a scene, which removes spatial information between each features.

\textbf{CNN derived descriptors for SLAM}
There has been significant interest in utilizing CNN-e objects instead of feature descriptors in SLAM~\cite{Quadslam,seman_map}, by engaging object detection algorithms like R-CNN~\cite{RCNN} and Mask R-CNN~\cite{maskrcnn} to extract objects from a scene.
Research by Sunderhauf \etal ~\cite{sunderhauf} uses features from the higher layers of the ConvNet hierarchy to encode semantic information about a scene, but due to the large size of extracted features, it is difficult to use it in real-time SLAM. 
Recent work by Wang \etal ~\cite{Saliency} uses salient regions for loop closure detection, where it extracts salient objects using CNN and uses them for saliency detection and re-identification in a scene. However, it does not take the spatial relationships of objects into account.

\textbf{Ensemble-based machine learning algorithms}
In machine learning, ensemble methods use multiple trained models to obtain better predictive performance. They combine multiple learned models into a single predictive model, in order to decrease variance and bias \cite{ensemble}.
Random Forests are a combination of tree predictors such that each tree uses input vectors that are sampled independently, with the same distribution for all trees in the forest.
Using a random selection of features and instances and multiple decision trees allow it to have low error rates and greater robustness with respect to noise \cite{randforest}.

\section{Proposed Methods}
\label{sec:proposedmethod}
In our proposed SymbioLCD, objects are first extracted using CNN, then they go through a filtering process based on semantics, size and placement.
Second, the spatial and semantic information received from the CNN gets projected onto a normalised plane to be scale-invariant. The semantic matching is performed on normalised object distances.
Third, the 2D spatial matching is performed using the Hausdorff distance with temporal similarity constraints.
Lastly, CNN-e information, e.g. semantic matches, Hausdorff distances and normalised distances, are combined with the vBoW similarity score, and the combined information is then used by the Ensemble-based LCD for predicting LCD candidates.
The overview of modularized components of the SymbioLCD is summarized in Figure \ref{fig:det_overview}. 
\subsection{Framework}
We propose SymbioSLAM as the framework for incorporating the SymbioLCD system. 
The SymbioSLAM framework was created using DynaSLAM \cite{dynaslam} as its base, and it was chosen because it incorporates ORB-SLAM2 \cite{ORBSLAM} that uses DBoW2 as its LCD module, and Mask R-CNN \cite{maskrcnn} for extracting objects from input images.
Components including, tracking, loop closing and frame drawing were modified to create the SymbioLCD system inside the SymbioSLAM framework.
\subsection{Object extraction using CNN}
\label{sec:objextraction}
The object extraction part of SymbioLCD was decoupled from the rest of processes, to make it CNN model agnostic. The CNN in this process is used only as an object-extractor and does not affect the ensemble-based LCD prediction process.
For object extraction, Mask R-CNN was pre-trained on the Microsoft COCO-dataset using ResNet50 as its backbone. After the objects are extracted from the CNN, the class label and bounding box get transferred to the filtering process. 
\subsection{Object filtering process}
\label{sec:objfiltering}
Once CNN completes its object extraction, the results are then filtered before they get transferred to the next process. 
Motivated by Bescos \etal~\cite{dynaslam} we use object labels to remove any moving objects, e.g. person, scooter, bicycle \& etc, and any objects which are larger than a certain threshold, e.g. a big table that overlaps other objects, from the list of extracted objects. 
By excluding moving objects it eliminates accidental loop closing on moving objects, and selecting a limited number of the largest objects in the scene helps it to cope better when repetitive objects are present in the scene. 



\subsection{Semantic matching in scale invariant normalised plane}
\label{sec:labelmatching}
The spatial information, i.e. the distances between each pair of objects, get normalised to make them scale-invariant, which helps with the semantic matching process. 
The normalisation of distances helps align class labels in a normalised plane, i.e. between 0 and 1, assisting them to find and match their correct nearest neighbour. The matching algorithm allows up to 40\% of misclassified or missing labels as long as other object labels and their position in the normalised plane matches the reference. This makes the label matching process much more robust, as it can account for detection errors that propagated from the CNN extraction process. 
Periodically, these vectors get assigned to a database, creating an online vocabulary of vectors to be used as references for finding loop closure candidates.

\subsection{Spatial matching using Hausdorff distance}

After the semantic matching is completed, object coordinates and their formation are analysed using Hausdorff distance. The normalised distance matching in the previous process is a 1D matching algorithm, i.e. comparing distances between a pair of objects, whereas the Hausdorff distance is a 2D matching algorithm using object coordinates. 

Euclidean distance is the most common way to measure dissimilarity between elements, however Euclidean distance does not take the spatial relations between all elements into account. Therefore SymbioLCD applies Hausdorff distance to compute the spatial relationships between all objects in the vector \cite{Hausdorff_eq}.

Given two vectors, the current vector containing the current objects' positions is written as
\begin{equation}
A_{c} = (a_1,...a_n),  
\end{equation}

and a reference vector containing a previous objects' positions is written as
\begin{equation}
B_{r} = (b_1,...b_n).  
\end{equation}

For every point in \textit{A\textsubscript{c}}, the distance to the nearest point in \textit{B\textsubscript{r}} is calculated, and the maximum value is assigned to \textit{h(A\textsubscript{c}, B\textsubscript{r})}. Likewise, for every point in \textit{B\textsubscript{r}}, the distance to the nearest point in \textit{A\textsubscript{c}} is calculated and assigned to \textit{h(B\textsubscript{r}, A\textsubscript{c})}. The maximum of \textit{h(A\textsubscript{c}, B\textsubscript{r})} and \textit{h(B\textsubscript{r}, A\textsubscript{c})} is the Hausdorff distance \cite{Hausdorff_eq}

\begin{equation}
H(A_{c},B_{r}) = \max( h(A_{c},B_{r}), h(B_{r},A_{c})),
\end{equation}

where
\begin{equation}
h(A_{c},B_{r}) = \max\limits_{a\epsilon A_{c}} \min\limits_{b\epsilon B_{r}} \|a-b\|.
\end{equation}

\subsection{Temporal similarity constraints}
\label{sec:tempsimilar}
Since adjacent images in the sequence look very similar to each other, we adjust the Hausdorff distance by adding the temporal similarity equation proposed by Zhang \etal~\cite{Temporal_similarity}

\begin{equation}
S^t(i,j) = exp(-\frac{\beta_sv^2_c}{f_c}(i-j)^2),
\end{equation}

\textit{i} and \textit{j} are the indices of two images being compared, \textit{v\textsubscript{c}} is the velocity of the camera, $\beta$\textsubscript{s} is a constant parameter and \textit{f\textsubscript{c}} is the frame rate.
To simplify, we've set \textit{v\textsubscript{c}} and \textit{f\textsubscript{c}} to be constant~\cite{compressed_holistic}, giving us

\begin{equation}
S^t(i,j) = exp(-\beta_s(i-j)^2),
\end{equation}

where
\begin{equation}
\beta_s \: \epsilon \: (0,1).
\end{equation}

Zhang \etal\ use this formula to aid their similarity equation, where they were looking to maximise the similarity score. However, in our case we are trying to find the minimum Hausdorff distance in a scene. Therefore we use the inverse of the function and negate the $\beta$\textsubscript{s}

\begin{equation}
S^t(i,j) = \frac{1}{exp(\beta_s(i-j)^2)}.
\end{equation}

By combining (8) with the Hausdorff distance measurement in (3), it could be rewritten as

\begin{equation}
H^t(i,j) = H(A_{i},B_{j}) + \alpha(S^t(i,j)),
\end{equation}
where $\alpha  \: \epsilon \: (0,100) $ is a parameter to control the weight of $S^t(i,j)$.
When image \textit{i} and \textit{j} are close, the value of $S^t(i,j)$ will be large, increasing the Hausdorff distance. If the image \textit{i} and \textit{j} are far apart, the value of $S^t(i,j)$ will become negligible, and won't affect the Hausdorff distance. With the temporal constraint, it becomes easier to distinguish and penalize frames that are very close, improving the overall accuracy of the LCD.

\subsection{LCD prediction using Random Forest}
\label{sec:lcdprediction}
The information from CNN-e objects, i.e. Matched labels, Hausdorff distance, Normalised distances, are concatenated with the vBoW similarity score and the combined information gets passed on to the ensemble-based LCD predictor. We use Random Forest as our ensemble-based LCD predictor, which contains an ensemble of Decision Trees. Random Forest was chosen, as it is fast, robust and less prone to overfitting when compared to other classical algorithms, such as SVM and Decision Tree \cite{randforest}. 

Given a vector $X= (X1,...,Xn)$ as input and a vector \textit{Y} representing the output, i.e. binary LCD classification, it assumes an unknown joint distribution $P_{XY}(X,Y)$ 
with  a loss function $L(Y,h(X))$. 
A classification function $h(x)$ can be defined as \cite{randforest}
\begin{equation}
h(x) = arg\max\limits_{y\epsilon n}P(Y=y|X=x),
\end{equation} 
where $n$ denotes a set of possible values of $y$.

An ensemble constructs $f$ in terms of a collection of base learners $h_1(x)...h_i(x)$, and the base learners are combined to give the ensemble prediction $f(x)$

\begin{equation}
f(x) = arg\max\limits_{y\epsilon n} \sum_{j=1}^{i} I(y=h_j(x)). 
\end{equation}
When the training set is small, Random Forest often outperforms Neural Networks, which tends to require large amounts of data to attain high accuracy. 




\section{Experiments}
\label{sec:experiments}
We carried out following experiments to evaluate our proposed methods. Section \ref{sec:datasets}  describes the datasets used in the experiments and evaluation parameters, Section \ref{sec:measuringsymbiosis} evaluates symbiosis between CNN-e objects information and vBoW features, Section \ref{sec:precision} compares the ensemble-based LCD predictor against widely used machine learning algorithms and Section \ref{sec:keyframe} compares the performance of LCD candidate detection against state-of-the-art SLAM systems. Experiments were performed on a PC with Intel i9 7900X CPU and GTX1080Ti GPU.

\begin{table*} [h!]
\centering
\caption{Precision and Recall Comparisons}
\label{tab:precision}
\begin{tabular}{ccccccccccccc} \toprule \midrule
      
    &\multicolumn{2}{c}{SVM-Linear}
    &\multicolumn{2}{c}{SVM-Polynomial}
    &\multicolumn{2}{c}{SVM-RBF}
    &\multicolumn{2}{c}{DecisionTree}
    &\multicolumn{2}{c}{SymbioLCD}
    &\multicolumn{2}{c}{NeuralNetwork}
   \\
      Dataset &Precision&Recall &Precision&Recall &Precision&Recall &Precision&Recall &Precision&Recall &Precision&Recall
      
      \\ \midrule
      fr2-desk       &80.77&64.62&66.29&90.77&90.91&83.33&100.00&75.38&92.31&100.00&76.92&83.33
      \\
      fr3-longoffice  &83.10&75.64&77.13&92.95&67.74&77.78&95.24&76.92&96.30&94.43&66.67&88.89
      \\
      lounge    &80.77&74.67&79.47&92.89&85.00&83.95&93.55&77.33&98.77&97.57&86.67&96.30   
      \\
      kitchen   &81.88&77.22&81.49&93.35&89.73&83.44&93.16&77.53&99.32&93.63&91.07&97.45
      \\
      garden     &84.14&77.59&84.03&94.34&89.04&86.28&94.90&79.01&99.54&95.13&90.20&97.79
      \\ \midrule 
      Average    &82.13&73.95&77.68&92.86&84.48&82.96&95.37&77.24&97.25&96.16&82.31&92.75
      \\
      \bottomrule

\end{tabular}
\end{table*}

\begin{table} [ht]
\caption{Parameters and Datasets}
 \begin{subtable}[h]{0.4\columnwidth}
 \scriptsize
 \centering
  \begin{tabular}{cc} \toprule \midrule
      Parameters & Value    \\ \midrule
      $\alpha$ & 100  \\
      $\beta_s$  & 1   \\
      Training  & 2453 \\
      Positive  & 300 \\
      Estimators            &100  \\
      Max features          &sqrt  \\
     \midrule \bottomrule
  \end{tabular}
  \caption{Parameters}
  \label{tab:parameters}
 \end{subtable}
 \hfill
 \begin{subtable}[h]{0.6\columnwidth}
 \centering
 \scriptsize
  \begin{tabular}{cccc} \toprule \midrule
      Dataset & Source & No. of  & Image   \\
      & & frames & Res.\\
      \midrule
      fr2-desk       & TUM  & 2965   &   640x480\\
      fr3-longoffice & TUM   & 2585  &   640x480\\
      lounge         & ours & 1841   &   640x480\\
      kitchen         & ours & 1998  &   640x480\\
      garden         & ours & 2148   &   640x480\\ \midrule \bottomrule
  \end{tabular}
  \caption{Datasets descriptions}
  \label{tab:dataset}
 \end{subtable}
\end{table}

\subsection{Datasets}
\label{sec:datasets}
Since SymbioLCD assumes a scene where common objects are present and that the camera re-visits the start of the trajectory to enable loop closure, two public datasets from TUM~\cite{TUM_data}, fr2-desk and fr3-long office, and three of our own datasets, lounge, kitchen and garden were used. 
Our datasets were taken using a camera with Sony IMX240 sensor, and represent a normal household area.

\begin{figure}[h]
\begin{center}
\includegraphics[width=0.8\linewidth]{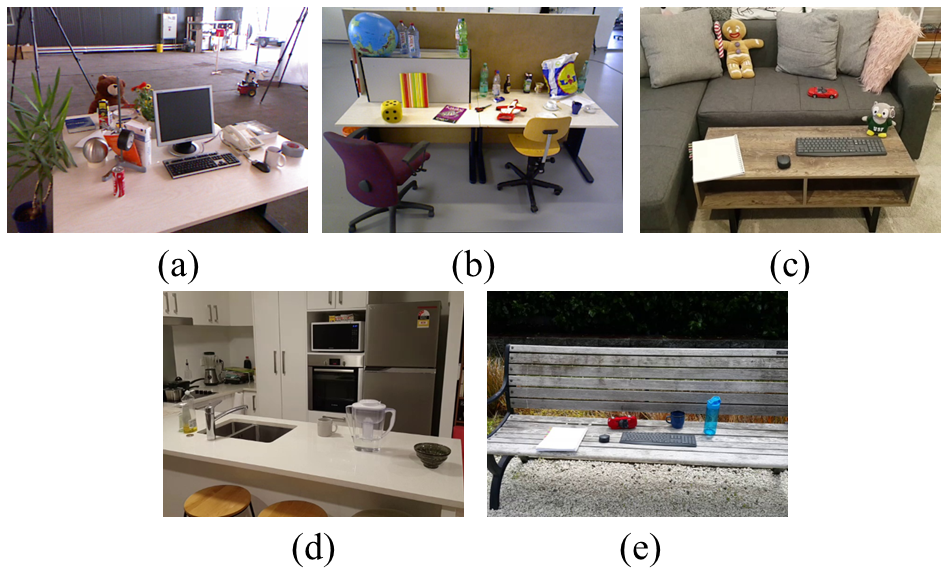}
\end{center}
   \caption{Evaluation datasets (a) fr2-desk (b) fr3-longoffice (c) lounge (d) kitchen (e) garden}
   \label{fig:datasets}
\end{figure}

Descriptions of the datasets are summarized in Table \ref{tab:dataset}, and Figure \ref{fig:datasets} shows sample images from each dataset. Table \ref{tab:parameters} displays parameters used for evaluation of ensemble-based LCD.


\subsection{Evaluating symbiosis - between CNN-e objects and vBoW}
\label{sec:measuringsymbiosis}
To show the extent of symbiosis between CNN-e objects information and vBoW features, the feature importance was measured on all inputs to the ensemble-based LCD, e.g. normalised distances, Hausdorff distances, matched class labels and vBoW similarity score. Also, an ablation study was performed to measure the difference between using both CNN-e objects and vBoW scores versus using one feature-set at a time, e.g. using just CNN-e objects or vBoW scores. The feature importance and ablation study are good measures of symbiosis, as it would indicate how much each feature-set would affect the ensemble-based LCD's prediction.

\begin{figure}[h]
\begin{center}
\includegraphics[width=0.9\linewidth]{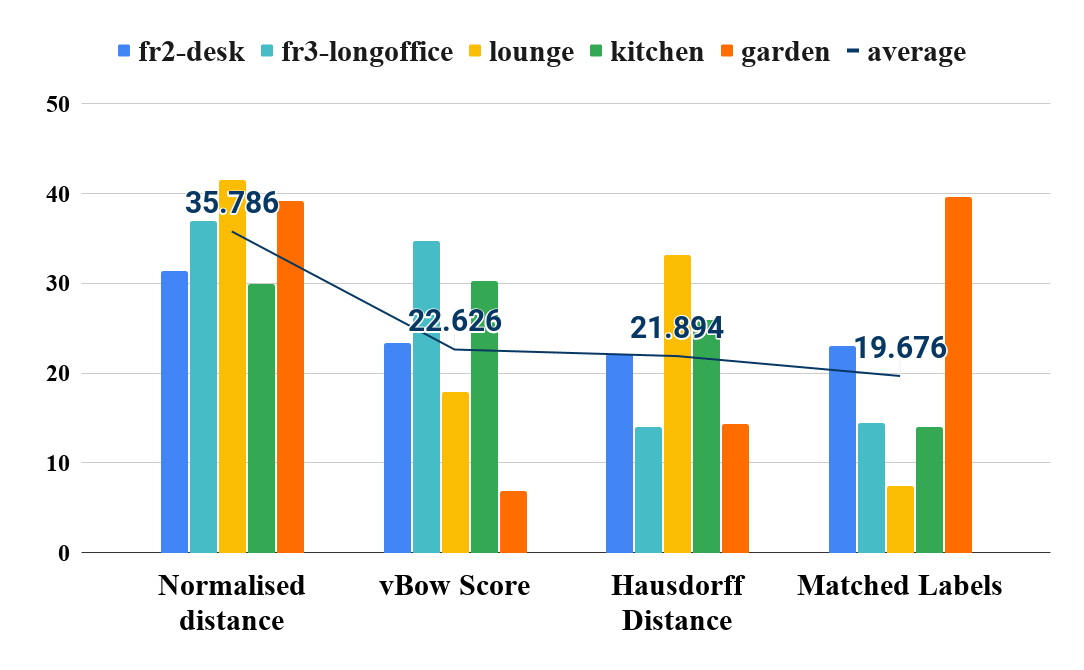}
\end{center}
   \caption{Feature importance between all features}
   \label{fig:featureimp}
\end{figure}

Figure \ref{fig:featureimp} shows that on average, 'Normalised distance' has the highest importance at 35.79\%, followed by 'vBoW score' at 22.63\%, 'Hausdorff distance' at 21.89\% and 'Matched labels' at 19.68\%. The results shows that normalised distances and vBoW score are the top two input features that influence the LCD prediction, with combined percentage of 58.41\%.

Further to that, we have performed an ablation study where we compare keyframe prediction results using both feature-sets, e.g. both CNN-e objects and vBoW score, against using one feature-set at a time, e.g using only CNN-e objects or VBoW score.
Table \ref{tab:ablation} shows that on average, using both feature-sets outperforms using just a single feature-set.
Compared to just using CNN-e objects, using both feature-sets provides an average improvement of 5.68\% in precision and 8.38\% in recall. Compared to using vBoW alone, it provides an average improvement of 53.36\% in precision and 50.95\% in recall. This demonstrates that there is a strong symbiotic relationship between CNN-e objects information and vBoW features, and using both feature-sets enables higher accuracy compared to using just a single feature-set.




\begin{table} [h!]
\centering
\caption{Ablation study - CNN-e vs. vBoW vs. Both}
\label{tab:ablation}
\begin{tabular}{ccccccc} \toprule \midrule
      
    &\multicolumn{2}{c}{CNN-e}
    &\multicolumn{2}{c}{vBoW}
    &\multicolumn{2}{c}{Both}

   \\
      Dataset &Preci.&Recall &Preci.&Recall &Preci.&Recall
      
      \\ \midrule
      fr2-desk       &90.07 &89.65 &45.69 &46.23 &92.31 &100.00
      \\
      fr3-longoffice  &90.82 &87.23 &46.86 &51.89 &96.30 &94.43
      \\
      lounge    &90.78 &88.08 &44.16 &47.11 &98.77 &97.57  
      \\
      kitchen   &92.21 &86.59 &46.99 &50.00 &99.32 &93.63
      \\
      garden     &93.94 &87.32 &35.71 &30.77 &99.54 &95.13
      \\ \midrule 
      Average    &91.56 &87.77 &43.88 &45.20 &97.25 &96.16
      \\
      \bottomrule

\end{tabular}
\end{table}






\subsection{Evaluating LCD keyframe prediction against other ML algorithms}
\label{sec:precision}
SymbioLCD was benchmarked against five other widely used machine learning algorithms, e.g. SVM with linear kernel, SVM with polynomial, SVM with Gaussian RBF, Decision Tree and densely connected Neural Networks (NN) with four layers\footnote{Due to the small amount of training data, an NN with 4x18 layers was selected as it performed the best compared to other deeper networks.}, for LCD keyframe prediction.
For this evaluation, each algorithm was tasked to predict keyframes where the loop closure has occurred, and their performance was measured using the precision and recall metrics.
The evaluation was performed 50 times to account for the non-deterministic nature of the algorithms.

%
The precision and recall metrics were defined as follows
\begin{equation}
Precision= \frac{TP}{TP+FP}  \;\;\; \& \;\;\; Recall= \frac{TP}{TP+FN},
\end{equation}
where TP refers to true positive, FP refers to false positive and FN for false negative.

Table \ref{tab:precision} presents the precision and recall values of SymbioLCD compared with other algorithms for each dataset. The result shows that on average, SymbioLCD achieved the highest precision and recall compared to other algorithms. 





Further to that, we have used Autorank \cite{autorank} to analyse the performance of SymbioSLAM against other ML algorithms compared on all datasets.
Autorank analyses the result to determine differences in the central tendency, e.g. mean rank (MR), median (MED) and median absolute deviation (MAD), with paired samples that are independent of each other to compare the mean performance of each algorithm. The analysis was also plotted using Critical Difference diagram \cite{cd_diagram} to visualise the performance difference between each algorithm.

\begin{figure}[h]
\begin{center}
\includegraphics[width=0.8\linewidth]{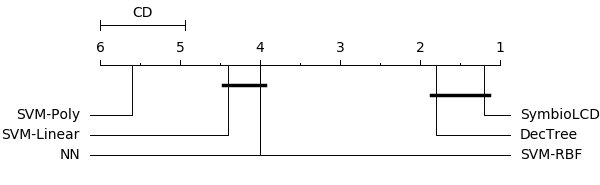}
\end{center}
  \caption{Critical Difference diagram (Precision) }
  \label{fig:autorank_pres}
\end{figure}

\begin{figure}[h]
\begin{center}
\includegraphics[width=0.8\linewidth]{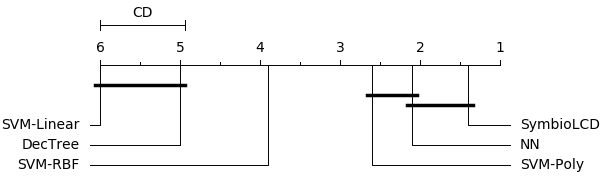}
\end{center}
  \caption{Critical Difference diagram (Recall) }
  \label{fig:autorank_rec}
\end{figure}

Table \ref{tab:stat_precision} and Figure \ref{fig:autorank_pres} shows that SymbioLCD performed best, i.e. ranked highest, compared to other ML algorithms in precision metric, and Table \ref{tab:stat_recall} and Figure \ref{fig:autorank_rec} shows that SymbioLCD performed best compared to other ML algorithms in recall metric on all datasets.

\begin{table}[h]
\centering
\caption{Autorank analysis (Precision)}
\label{tab:stat_precision}
\begin{tabular}{lrlllll}
\toprule
{} &    MR &   MED &   MAD &              CI & $\delta$ &   Mag. \\
\midrule
SymbioLCD  & 1.20 & 0.98 & 0.011 &  [0.96, 0.99] &         0 &  neglig. \\
DecTree    & 1.80 & 0.94 & 0.020 &  [0.93, 0.95] &     0.28 &       small \\
SVM-RBF    & 4.00 & 0.89 & 0.028 &  [0.85, 0.89] &     1.00 &       large \\
NeuralNets         & 4.00 & 0.86 & 0.065 &  [0.76, 0.90] &     1.00 &       large \\
SVM-Linear & 4.40 & 0.81 & 0.016 &  [0.80, 0.83] &     1.00 &       large \\
SVM-Poly   & 5.60 & 0.79 & 0.035 &  [0.77, 0.81] &     1.00 &       large \\
\bottomrule
\end{tabular}
\end{table}

\begin{table}[h]
\centering
\caption{Autorank analysis (Recall)}
\label{tab:stat_recall}
\begin{tabular}{lrlllll}
\toprule
{} &    MR &   MED &   MAD &              CI & $\delta$ &   Mag. \\
\midrule
SymbioLCD  & 1.40 & 0.96 & 0.037 &  [0.95, 0.98] &         0 &  neglig. \\
NN         & 2.10 & 0.96 & 0.022 &  [0.88, 0.97] &     0.32 &       small \\
SVM-Poly   & 2.60 & 0.92 & 0.006 &  [0.92, 0.93] &     0.92 &       large \\
SVM-RBF    & 3.90 & 0.83 & 0.008 &  [0.83, 0.84] &     1.00 &       large \\
DecTree    & 5.00 & 0.77 & 0.006 &  [0.76, 0.78] &     1.00 &       large \\
SVM-Linear & 6.00 & 0.75 & 0.023 &  [0.74, 0.77] &     1.00 &       large \\
\bottomrule
\end{tabular}
\end{table}

\subsection{Evaluating LCD keyframe prediction against vBoW-based SLAM systems}
\label{sec:keyframe}


To evaluate the performance of SymbioLCD, it was bench-marked against the state-of-the-art ORB-SLAM2 and DynaSLAM, which use vBoW for LCD. 
They were chosen because all three systems - our SymbioSLAM, ORB-SLAM2 and DynaSLAM, share the same vBoW algorithm and keyframe insertion algorithm, making them an ideal baseline to compare against our performance.

For this evaluation, we recorded the keyframe number when the loop closure candidate was found to benchmark its performance. As proposed by the author of ORB-SLAM2~\cite{ORBSLAM}, the evaluation was performed 5 times and the median value was recorded to account for the non-deterministic nature of the system.

\begin{figure*}[h]
\begin{center}
\includegraphics[width=0.42\linewidth]{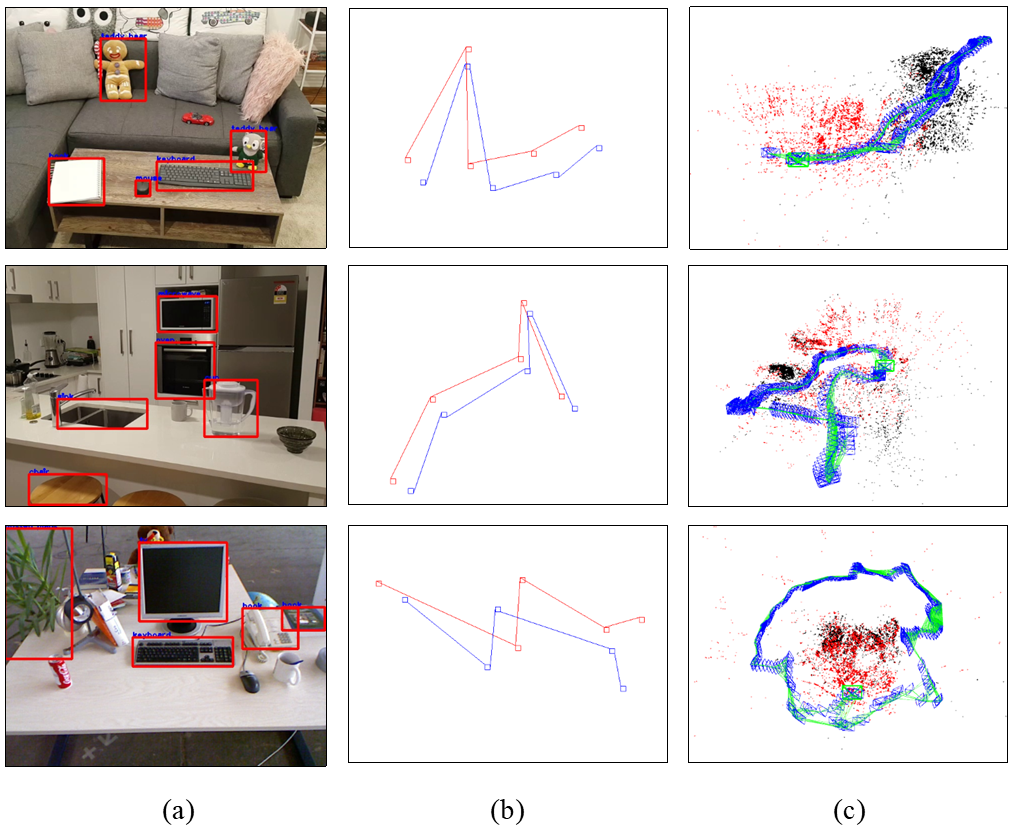}
\end{center}
   \caption{Results of lounge, kitchen and fr2-desk dataset. (a) live image with CNN-e objects (b) matched loop closure detection candidate. Red line represents the object-matrix from the current frame and blue represents the reference it matched on. (c) trajectory after the loop closure}
\label{fig:keyframe}
\end{figure*}

Table \ref{tab:keyframe} shows loop closure detected keyframes in each dataset and Figure \ref{fig:keyframe} shows samples of the live image, matched loop closure detection candidate and trajectories after the loop closure. 

\begin{table} [h]
\centering
\caption{Comparisons of loop closure detected keyframe.}
\label{tab:keyframe}
\begin{tabular}{ccccc} \toprule \midrule
      Dataset &\vtop{\hbox{\strut Symbio}\hbox{\strut LCD}\hbox{\strut (kf)}}& \vtop{\hbox{\strut ORB}\hbox{\strut SLAM2}\hbox{\strut (kf)}} &
      \vtop{\hbox{\strut Dyna}\hbox{\strut SLAM}\hbox{\strut (kf)}} & 
      \vtop{\hbox{\strut Std.}\hbox{\strut Dev.}}
      \\ \midrule
     fr2-desk & 388 & 393 & 397 & 1.92\\
     fr3-longoffice & 314 & 345 & 349 & 2.38\\
     lounge         & 284 & 301 & 303 & 2.45\\
     kitchen         & 392 & 410 & 416 & 1.22\\
     garden       & 450 & 454 & 459 & 2.07\\ 
\midrule \bottomrule

\end{tabular}
\end{table}

The result in Table \ref{tab:keyframe} shows that SymbioLCD has consistently outperformed both state-of-the-art systems in all evaluated datasets, exceeding ORB-SLAM2 by an average of 4.52\% (15 keyframes) and DynaSLAM by 5.65\% (19.2 keyframes).
Since SymbioLCD focuses on spatial relations of selected objects in the scene, it was able to acquire loop closure candidates early compared to ORB-SLAM2 and DynaSLAM, which only rely on vBoW. This evaluation demonstrates that having a spatial-awareness of the scene helps to detect loop closure candidates earlier than the state-of-the-art. 

\section{Conclusion and Future Work}
\label{sec:conclusion}
We presented our novel SymbioLCD with an ensemble-based LCD that uses both CNN-e object information and vBoW features. We showed that there is strong symbiosis between CNN-e objects and vBoW features in SymbioLCD's decision making process. Our proposed method outperformed other ML algorithms on all datasets in precision and recall evaluation, and it was able to locate loop closure candidates earlier than state-of-the-art SLAM algorithms.
SymbioLCD requires multiple static objects in a scene to be most efficient. As future research, we aim to extend SymbioLCD to work with both static and dynamic objects.

{\small
\bibliographystyle{IEEEtran}
\bibliography{egbib}
}

\end{document}